\documentclass[conference]{IEEEtran}
\IEEEoverridecommandlockouts
\usepackage{cite}
\usepackage{amsmath,amssymb}
\usepackage{graphicx}
\usepackage{booktabs}
\usepackage{tikz}
\usetikzlibrary{arrows.meta,positioning,fit,backgrounds}
\usepackage[hidelinks,breaklinks]{hyperref}
\usepackage[capitalize]{cleveref}
\crefname{table}{Tab.}{Tabs.}\crefname{figure}{Fig.}{Figs.}%
\crefname{section}{Sec.}{Secs.}\crefname{equation}{Eq.}{Eqs.}

\newcommand{\Dtwo}{SAL}
\newcommand{\Pmtl}{P_{\mathrm{MTL}}}
\newcommand{\arxBaseline}{0.45}

\newcommand{\arxVApoolP}{0.005}
\newcommand{\arxTwoBBPMTL}{1.6669}

\newcommand{\arxTwoBBEXPR}{0.4698}
\newcommand{\arxTwoBBAU}{0.5564}
\newcommand{\arxThreeBBPMTL}{1.6949}

\newcommand{\arxThreeBBEXPR}{0.4978}

\newcommand{\arxThreeBBcalAU}{0.5699}
\newcommand{\arxFourBBcalPMTL}{1.7099}
\newcommand{\arxFourBBcalVA}{0.6407}

\newcommand{\arxVApoolPMTL}{1.7259}
\newcommand{\arxVApoolVA}{0.6567}

\newcommand{\arxFfsfm}{0.4461}
\newcommand{\arxFmae}{0.4416}
\newcommand{\arxFloraan}{0.4762}

\newcommand{\arxFemo}{0.4540}

\newcommand{\arxRcurriccurric}{0.98}
\newcommand{\arxRloralora}{0.91}
\newcommand{\arxRloracross}{0.75}
\newcommand{\arxRfsfmmae}{0.81}

\begin{document}
\title{Strength-Parity Ensembling with Parameter-Isolated Experts for Multi-Task Affect Recognition}
\author{\IEEEauthorblockN{Tung Hung Bui, Hong Hai Nguyen}%
\IEEEauthorblockA{Dept. of AI, FPT University}
\and
\IEEEauthorblockN{Van Thong Huynh}%
\IEEEauthorblockA{Fac. of CSE, HCMC University of Technology - VNUHCM}
}
\maketitle

\begin{abstract}
Leading entries on the multi-task track of the 11th ABAW challenge rely on heavy ensembling, yet which
member is worth adding to an already strong ensemble is rarely made explicit. We study this question for
joint valence--arousal estimation, 8-way expression recognition, and 12-way action-unit detection
from a single unconstrained face, under partial, long-tailed labels and a rule that forbids pretraining on
Aff-Wild2. Building on a shared affect-latent that marginalizes the missing labels across two
affect-supervised backbones, we propose a strength-parity rule: an added member lowers the
ensemble error only when it is both decorrelated from the current members and a near-peer of them in
individual accuracy. The rule exposes a concrete obstacle, as on a single backbone re-seeding and even
distinct fine-tuning curricula re-converge to a prediction correlation of \arxRcurriccurric{} and add no
diversity. Parameter-isolation removes it: confining each adaptation to a disjoint low-rank subspace of a
shared backbone yields experts that stay decorrelated at \arxRloralora{} while remaining near-peers, the
strongest of them an AffectNet-adapted expert. The resulting system raises the overall validation score to
\arxThreeBBPMTL{}, against the organizers' ConvNeXt-with-MixAugment baseline of
\arxBaseline; with per-AU calibration and by pooling the shared-latent heads'
valence--arousal byproduct as a further near-peer, the strongest configuration reaches \arxVApoolPMTL{}.

\end{abstract}

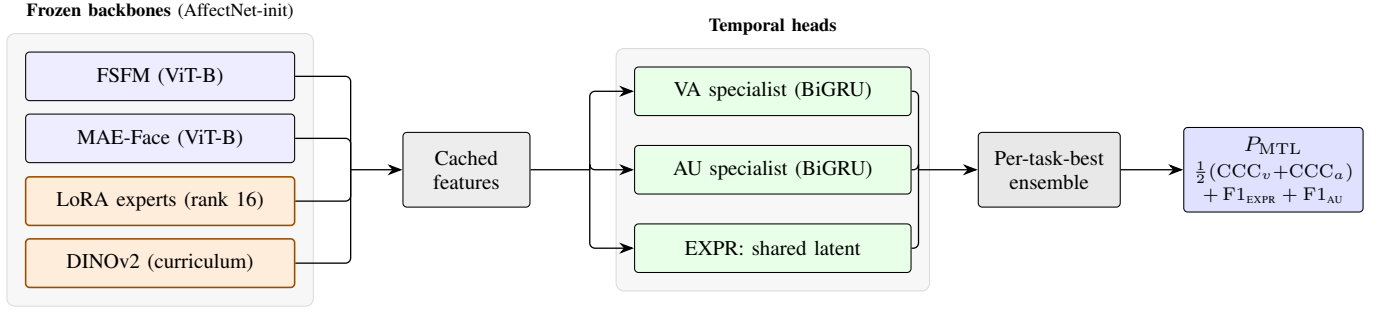
\begin{figure*}[t]
\centering
\resizebox{0.99\textwidth}{!}{%
\begin{tikzpicture}[font=\footnotesize,
  bb/.style={draw, rounded corners=1.5pt, fill=blue!7, align=center, minimum width=36mm, minimum height=6.5mm, inner sep=2pt},
  ours/.style={bb, draw=orange!60!black, fill=orange!14, semithick},
  feat/.style={draw, rounded corners=1.5pt, fill=black!7, align=center, minimum width=17mm, minimum height=10mm},
  head/.style={draw, rounded corners=1.5pt, fill=green!9, align=center, minimum width=37mm, minimum height=6.5mm, inner sep=2pt},
  asm/.style={draw, rounded corners=1.5pt, fill=black!9, align=center, minimum width=19mm, minimum height=10mm},
  res/.style={draw, rounded corners=1.5pt, fill=blue!12, align=center, minimum width=24mm, minimum height=10mm},
  grp/.style={rounded corners=3pt, draw=black!14, fill=black!3, inner sep=2.4mm},
  ar/.style={-{Stealth[length=2mm]}, line width=0.5pt, rounded corners=2pt}]
  \node[bb]   (fsfm) at (0,1.25)  {FSFM (ViT-B)};
  \node[bb]   (mae)  at (0,0.42)  {MAE-Face (ViT-B)};
  \node[ours] (lora) at (0,-0.42) {LoRA experts (rank 16)};
  \node[ours] (dino) at (0,-1.25) {DINOv2 (curriculum)};
  \node[feat] (feat) at (4.1,0) {Cached\\ features};
  \node[head] (va)   at (8.2,1.05) {VA specialist (BiGRU)};
  \node[head] (au)   at (8.2,0)    {AU specialist (BiGRU)};
  \node[head] (expr) at (8.2,-1.05){EXPR: shared latent};
  \node[asm]  (asm)  at (11.9,0) {Per-task-best\\ ensemble};
  \node[res]  (out)  at (14.9,0) {$\Pmtl$\\[1pt] {\scriptsize$\tfrac{1}{2}(\mathrm{CCC}_v{+}\mathrm{CCC}_a)$}\\ {\scriptsize$+\,\mathrm{F1}_{\textsc{expr}}+\mathrm{F1}_{\textsc{au}}$}};
  \begin{scope}[on background layer]
    \node[grp, fit=(fsfm)(mae)(lora)(dino)] (gbb) {};
    \node[grp, fit=(va)(au)(expr)] (ghead) {};
  \end{scope}
  \node[font=\scriptsize\bfseries, anchor=south, yshift=0.6mm] at (gbb.north) {Frozen backbones {\normalfont(AffectNet-init)}};
  \node[font=\scriptsize\bfseries, anchor=south, yshift=0.6mm] at (ghead.north) {Temporal heads};
  \coordinate (m1) at (2.55,0);  \coordinate (m2) at (5.75,0);  \coordinate (m3) at (10.15,0);
  \foreach \b in {fsfm,mae,lora,dino} \draw[line width=0.5pt, rounded corners=2pt] (\b.east) -| (m1);
  \draw[ar] (m1) -- (feat.west);
  \draw[line width=0.5pt] (feat.east) -- (m2);
  \foreach \h in {va,au,expr} \draw[ar] (m2) |- (\h.west);
  \foreach \h in {va,au,expr} \draw[line width=0.5pt, rounded corners=2pt] (\h.east) -| (m3);
  \draw[ar] (m3) -- (asm.west);
  \draw[ar] (asm.east) -- (out.west);
\end{tikzpicture}}
\caption{System overview. Frozen affect-supervised backbones and the added members (highlighted:
parameter-isolated LoRA experts and a curriculum near-peer) supply cached per-frame features to BiGRU
specialists (valence--arousal, action units) and a shared affect-latent (expression), whose per-task-best
members are averaged. The shared latent and two-backbone base follow companion work~\cite{companion}.}
\label{fig:system}
\end{figure*}

\section{Introduction}
\label{sec:intro}
Recognizing emotion from an in-the-wild face is not a single problem but three that compete with one another.
The 11th Affective Behavior Analysis in-the-wild (ABAW) challenge~\cite{kollias2024abaw7,kollias2026affect}
makes this tension explicit in its multi-task learning (MTL) track: from a single cropped face on the
s-Aff-Wild2 corpus, a model must place the affect on the continuous valence--arousal circumplex, name one
of eight categorical expressions, and flag which of twelve facial action units are active. The three
label families are annotated on overlapping but unequal frame subsets, are individually long-tailed, and
the challenge forbids any use of Aff-Wild2 for pretraining. The ABAW series~\cite{kollias2019deep,kollias2022abaw,kollias2023abaw,kollias20246th,kollias2024abaw7,kollias2026affect}
has driven progress on this corpus, and leading MTL entries share a recipe: a face-pretrained transformer
backbone, task-specific heads, and heavy ensembling~\cite{abaw2022augraph,abaw2024auassisted,savchenko2024hsemotion},
with the current published validation figure set by the Progressive Learning system~\cite{abaw2024progressive}. The recurring difficulty is not the individual tasks
but their partial and imbalanced supervision; the standard responses mask the missing labels in the loss or
impute them with a teacher~\cite{wang2021meanteacher,gera2022ssmfar}.

A base system for this track already exists, and it takes a different route to the missing labels. Companion
work~\cite{companion} casts the partial-label problem as marginalization over a shared affect-latent
(\Dtwo): a small variational bottleneck mediates the three task decoders, so that a frame carrying only one
label still shapes the representation the other two rely on~\cite{kingma2014semi}, and pairs the latent with
two affect-supervised face backbones whose per-task predictions are averaged. We adopt that system as our
starting point rather than revisiting it.

Ensembling and its dependence on member diversity are long understood~\cite{krogh1995ensembles,brown2005diversity,lakshminarayanan2017deep},
but the ABAW literature typically diversifies by backbone or by seed and reports the pooled number. Our
question is narrower and operational: given this fixed strong ensemble, we ask when adding one more member
helps, and what property that member must possess.

The answer, our core contribution, is a compact admission criterion. An added member helps only when it
satisfies two conditions simultaneously: it must disagree with the current ensemble (be decorrelated) and be
about as accurate as the members already in it (a near-peer). We call this criterion strength-parity. Stated
in the abstract, it is nearly self-evident; what is not, and what matters in practice, is which of the
inexpensive ways of producing a ``new'' model in fact satisfy it. Re-seeding, and even running a different supervised curriculum on the same backbone, yield nearly the
same predictions; distinct full fine-tuning curricula on one backbone correlate at \arxRcurriccurric{},
and therefore add nothing to average over. Decorrelation is a property of the pretrained feature space
rather than of the path taken to fine-tune it.

This diagnosis is also constructive: if unconstrained fine-tuning re-converges, constraining it should not.
We adapt a single shared backbone to different affect corpora through disjoint low-rank subspaces,
in the manner of LoRA~\cite{hu2022lora} but repurposed from a parameter-efficient adapter into a diversity
mechanism~\cite{wang2023loraens}, and the resulting experts stay decorrelated, at a correlation of
\arxRloralora{} among experts on one backbone, while each remains a near-peer. An AffectNet-adapted low-rank
expert is the strongest single expression member in the pool, at macro-F1 \arxFloraan{}, above either base
backbone, and adding it raises the ensemble expression score from \arxTwoBBEXPR{} to \arxThreeBBEXPR{} and
the overall validation score to \arxThreeBBPMTL{}. This supplies the mechanism the strength-parity rule
requires: a means of producing decorrelated near-peers from a single backbone, without a second pretrained
one. In summary, our contributions are as follows
\begin{itemize}
\item A strength-parity rule for ensembling partially-labeled affect models: a member helps only if it is
  both decorrelated from the ensemble and an individual near-peer, which we use as a pre-ensemble admission
  test.
\item A mechanism that supplies such members: low-rank adaptation of a single backbone yields decorrelated
  near-peer experts where full fine-tuning re-converges, which explains why cheaper diversity sources fail.
\item Results on the 11th ABAW MTL track: the system reaches \arxThreeBBPMTL{} on validation and
  \arxVApoolPMTL{} in its strongest configuration, against a baseline of \arxBaseline{}, with every gain
  sized by a paired video-level bootstrap.
\end{itemize}

\section{Method}
\label{sec:method}
\Cref{fig:system} gives an overview of the full pipeline; we first summarize the base recognition system we
build on, then state the rule that governs which additional members improve it and the mechanism that
produces such members. The base system's design, ablations, and justification belong to the companion
work~\cite{companion} and are not claimed here.

\subsection{The shared affect-latent base system}
\label{sec:base}
Backbones and features. Two ViT-B/16 face backbones supply complementary representations: one is
self-supervised on VGGFace2 in the FSFM manner~\cite{wang2025fsfm}, the other masked-autoencoded on a large
face corpus in the MAE-Face manner~\cite{ma2023maeface}. Each passes through the same three stages:
categorical fine-tuning on AffectNet~\cite{mollahosseini2019affectnet} to make the features affect-aware, a
short per-frame adaptation on s-Aff-Wild2 with the top blocks unfrozen, and a freeze, after which the network
is a fixed extractor whose per-frame 768-dimensional embeddings are cached once. Every temporal model then
reads these cached features instead of back-propagating into the backbone, which is what makes an individual
member cheap enough to train and to ensemble at the scale the rest of the paper relies on.

Temporal heads and the shared latent. Over a window of cached features, a bidirectional GRU produces
per-frame states that feed three task decoders. Valence and arousal are regressed by a dedicated head under
a concordance loss, and the twelve action units by a multi-label head under binary cross-entropy; each loss
is evaluated only on the frames that carry the corresponding label, so a partially annotated frame still
trains the tasks it can. Expression is decoded through the shared affect-latent \Dtwo{}: a variational
bottleneck mediates all three decoders and is trained by a masked evidence lower
bound~\cite{companion,kingma2014semi} that marginalizes each frame's absent labels, so that a frame carrying
only action-unit labels still updates the latent the expression decoder reads. Routing expression through
this shared latent, rather than a task-specific head, yields per-backbone macro-F1 of \arxFmae{}--\arxFfsfm{}
and outperforms a matched masked-loss joint model.

The two backbones are individually near-peers and make different errors, so averaging their per-task
predictions improves every task; assembled per-task-best, the base system reaches \arxTwoBBPMTL{} on
validation. Its expression score, however, rests on only two shared-latent members, and an average is no
better than its members are diverse. The remainder of this section adds a member that improves it.

\subsection{The strength-parity rule}
\label{sec:rule}
Let the current ensemble average the per-frame posteriors of its $M$ members,
\begin{equation}
\label{eq:avg}
\bar{p} = \frac{1}{M}\sum_{i=1}^{M} p_{m_i},
\end{equation}
and consider adding one more. The averaged error of such an ensemble follows the classic
accuracy--diversity decomposition~\cite{krogh1995ensembles},
\begin{equation}
\label{eq:ens}
\mathbb{E}\,\lVert\bar{e}\rVert^2 = \varepsilon^2\!\left(\bar{\rho} + \frac{1-\bar{\rho}}{M}\right),
\end{equation}
where $\varepsilon^2$ is the mean squared error of the members and $\bar{\rho}$ their mean pairwise
correlation. The average therefore improves only when the new member lowers the correlation without raising
the error level: a candidate that duplicates the ensemble leaves the correlation untouched and contributes
nothing, however accurate, whereas a decorrelated but weak candidate raises the error level faster than it
lowers the correlation. The useful region is the intersection of the two, decorrelated and near-peer, and we
treat membership in it, judged from the correlation of a candidate's posteriors with the ensemble and from
its own task metric, as a pre-ensemble admission test rather than a property discovered after pooling.

The two conditions are genuinely separable: a re-seeded head on one backbone is strong but redundant,
while a general-purpose backbone is decorrelated but too weak for affect. A useful member must satisfy both
conditions, and the rest of this section builds such members, first through a supervised curriculum and
then, more cheaply, through parameter-isolation.

\subsection{A curriculum-built near-peer}
\label{sec:curriculum}
The natural source of decorrelation is a third backbone with a different pretraining, and the obvious
candidate is DINOv2~\cite{oquab2024dinov2}, whose general self-supervised features are the most
decorrelated from our face backbones. General pretraining alone is too weak for affect, so we give it a
supervised curriculum. Adapting DINOv2 through AffectNet and then EmotioNet~\cite{benitezquiroz2016emotionet} action-unit supervision before
the s-Aff-Wild2 stage turns it into an overall per-frame near-peer of the face backbones, and its
shared-latent expression member reaches macro-F1 \arxFemo{} while staying decorrelated from them, a
member that satisfies both conditions. The curriculum is chosen to match the in-the-wild target distribution
rather than to maximize accuracy on any single source corpus.

\subsection{Parameter-isolated experts}
\label{sec:lora}
The curriculum gives one near-peer, but each new one costs a full pretraining chain, and the deeper
problem remains: several full fine-tuning curricula on the same backbone re-converge to nearly the same
predictions, correlated at \arxRcurriccurric{}, so they cannot be pooled with one another. Unconstrained
fine-tuning erases the diversity we set out to create.

Constraining the adaptation preserves it: we freeze a shared near-peer backbone and adapt it to different
affect corpora through disjoint low-rank residual subspaces~\cite{hu2022lora}, one expert per corpus. Each
expert replaces a frozen projection weight $W_0$ by
\begin{equation}
\label{eq:lora}
W = W_0 + \tfrac{\alpha}{r}\,BA, \qquad B\in\mathbb{R}^{d\times r},\; A\in\mathbb{R}^{r\times k},
\end{equation}
with rank 16 and scaling factor 32, on the attention and MLP projections; the shared weight stays frozen,
and only the low-rank factors differ between experts. Because the experts occupy different low-rank subspaces
of the same frozen features, they specialize without collapsing onto one another, and two comparisons make
this concrete. On a single backbone, isolated experts sit at a prediction correlation of \arxRloralora{},
below the \arxRcurriccurric{} to which full fine-tuning re-converges, so the constraint prevents
same-backbone adaptations from collapsing to a single model. More importantly for our purpose, the strongest
expert is genuinely decorrelated from the two base backbones it joins, at \arxRloracross{}, comparable to the
\arxRfsfmmae{} between the two backbones themselves. Strength is preserved as well: the AffectNet-adapted
expert reaches expression macro-F1 \arxFloraan{}, above both base backbones and above the curriculum member,
and is the strongest single expression member in the pool. Parameter-isolation thus supplies what the
strength-parity rule requires, decorrelated near-peers produced from a single backbone, where full
fine-tuning could yield only decorrelated-or-strong, never both.

\subsection{Assembly and calibration}
\label{sec:assembly}
The final prediction is assembled per task from the members that are strongest on that task, all averaged at
the posterior level. Valence--arousal and action units use the two-backbone specialist averages of the base
system; expression averages the two shared-latent base members with the added parameter-isolated experts.
For action units we additionally fit a per-unit decision threshold that maximizes each unit's F1 on
validation. This calibration is fit on validation and therefore optimistic on held-out data; a video-level
cross-fit of the same threshold recovers only about a third of its apparent gain, so we
report the calibrated and uncalibrated numbers separately and treat the uncalibrated \arxThreeBBPMTL{} as
the robust number.

\section{Experiments}
\label{sec:exp}

\subsection{Evaluation protocol}
\label{sec:protocol}
The track scores a submission by summing the three tasks' native metrics,
\begin{equation}
\label{eq:pmtl}
\Pmtl = \tfrac{\mathrm{CCC}_{v}+\mathrm{CCC}_{a}}{2}
      + \tfrac{1}{8}\!\sum_{c=1}^{8}\mathrm{F1}^{\mathrm{expr}}_{c}
      + \tfrac{1}{12}\!\sum_{k=1}^{12}\mathrm{F1}^{\mathrm{au}}_{k}\;\in[0,3],
\end{equation}
the mean valence--arousal concordance plus the macro-averaged expression and action-unit F1.
We use the official s-Aff-Wild2 MTL partition~\cite{zafeiriou2017aff,kollias2019deep,kollias2019expression,kollias2019face,kollias2020analysing,kollias2021affect,kollias2021analysing,kollias2021distribution,kollias2022abaw,kollias2023abaw,kollias2023abaw2,kollias20246th,kollias2024abaw7,kollias2024distribution,kollias2024behaviour4all,kollias2025advancements,kollias2025emotions,kollias2026affect},
which is video-disjoint between training and validation, and
report on the validation split. All model
selection (hyperparameters, early stopping, member admission, and thresholds) uses only the training
and validation splits. Because the effective sample size at this granularity is the number
of videos rather than frames, we size every reported difference with a paired bootstrap over the fifty
validation videos and report the one-sided p-value; a gap that falls inside this noise is reported as
positive-within-noise rather than as a significant gain.

\begin{table*}[t]
\centering
\caption{Validation ladder on s-Aff-Wild2 MTL; each row adds one component. The shared-latent two-backbone
base follows companion work~\cite{companion}; rows below are contributed here. Per-AU calibration is
dev-fit. Best $\Pmtl$ in bold.}
\label{tab:challenge}
\begin{tabular}{@{}lcccc@{}}
\toprule
System & VA & \textsc{expr} & \textsc{au} & $P_{\mathrm{MTL}}$ \\
\midrule
Organizer baseline (ConvNeXt, MixAugment)~\cite{kollias2026affect} & -- & -- & -- & 0.45 \\
\midrule
Two backbones (FSFM + MAE-Face), \Dtwo{} \textsc{expr} & 0.6407 & 0.4698 & 0.5564 & 1.6669 \\
\;+ AffectNet-LoRA expert (\Dtwo{} \textsc{expr}) & 0.6407 & 0.4978 & 0.5564 & 1.6949 \\
\;+ per-AU threshold calibration & 0.6407 & 0.4978 & 0.5699 & 1.7084 \\
\;+ FSFM-LoRA expert & 0.6407 & 0.4993 & 0.5699 & 1.7099 \\
\;+ MAE-LoRA expert & 0.6407 & 0.4970 & 0.5699 & 1.7076 \\
\;+ pool \Dtwo{} members' VA into VA ensemble & 0.6567 & 0.4993 & 0.5699 & \textbf{1.7259} \\
\bottomrule
\end{tabular}

\end{table*}

Implementation details. The two ViT-B/16 backbones are AffectNet-initialized and adapted per-frame on
s-Aff-Wild2 with the top 40\% of blocks unfrozen, for three epochs at batch size 64 and learning rate
$5\times10^{-4}$. Low-rank experts use rank 16, scaling factor 32, and dropout 0.1 on the attention and MLP
projections. All temporal heads are bidirectional GRUs with hidden size 256, two layers, and dropout 0.2,
applied over windows of 64 frames at stride 32 and trained for up to 25 epochs at learning rate $10^{-3}$;
the shared affect-latent uses a 96-dimensional bottleneck with a KL weight of 0.05 annealed over the first
eight epochs. Each member is a seed average, ten seeds for the primary shared-latent member and six for the
experts and specialists.

\subsection{Main result}
\label{sec:main}
\Cref{tab:challenge} reports the validation cumulative results, on which the two-backbone shared-latent base system
scores \arxTwoBBPMTL{}. Adding the AffectNet parameter-isolated expert to the expression ensemble raises expression
macro-F1 from \arxTwoBBEXPR{} to \arxThreeBBEXPR{} and the total to \arxThreeBBPMTL{}, our robust primary 
result; the paired bootstrap places the added expert above the two-backbone expression ensemble by a
positive-within-noise margin at 50 videos. Per-AU calibration adds a further dev-fit gain on action
units, from \arxTwoBBAU{} to \arxThreeBBcalAU{}, and a second low-rank expert brings the system to
\arxFourBBcalPMTL{}, the point of diminishing returns for expression members. The diversity axis offers one
further member: the shared-latent members also produce a valence--arousal prediction as a byproduct, and
pooling it into the valence--arousal ensemble, alongside the two dedicated specialists, forms a further
decorrelated near-peer. It raises the mean CCC from \arxFourBBcalVA{} to
\arxVApoolVA{}, the one member gain to clear significance under the paired video bootstrap, at a p-value of
\arxVApoolP{}, and lifts the overall score to \arxVApoolPMTL{}, the strongest configuration on validation.

\subsection{The strength and diversity axes}
\label{sec:strength}
\begin{figure}[t]
\centering
\includegraphics[width=0.9\linewidth]{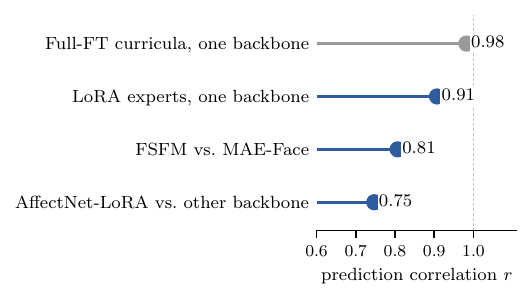}
\caption{Prediction correlation between expression members; lower correlation is more fusable. Full
fine-tuning of one backbone re-converges at \arxRcurriccurric{}, whereas parameter-isolation and a second
backbone stay decorrelated.}
\label{fig:decorr}
\end{figure}

\Cref{tab:members} scores each expression member on its own, and \cref{tab:decorr} reports the pairwise
prediction correlation. Together they place the admitted members in the useful region of both axes: the
two base members, the FSFM-LoRA expert, and the curriculum near-peer sit at macro-F1 \arxFmae{}--\arxFemo{},
the AffectNet-LoRA expert leads at \arxFloraan{}, and all of them are decorrelated enough to average well. The
correlation table summarizes the mechanism behind parameter-isolation in a single line, plotted in
\cref{fig:decorr}: full fine-tuning curricula on one backbone sit at a prediction correlation of
\arxRcurriccurric{}, low-rank experts on the same backbone at \arxRloralora{}, and the strongest expert
against a different backbone at \arxRloracross{}, comparable to the \arxRfsfmmae{} between the two base
backbones themselves.

\begin{table}[t]
\centering
\caption{Expression member strength (macro-F1 alone) for admitted members.}
\label{tab:members}
\footnotesize\setlength{\tabcolsep}{5pt}\begin{tabular}{@{}llc@{}}
\toprule
EXPR member & role & macro-F1 \\
\midrule
FSFM \Dtwo{} & base member & 0.4461 \\
MAE-Face \Dtwo{} & base member & 0.4416 \\
AffectNet-LoRA & isolated expert & 0.4762 \\
FSFM-LoRA & isolated expert & 0.4451 \\
DINOv2 curriculum & curriculum near-peer & 0.4540 \\
\bottomrule
\end{tabular}

\end{table}

\begin{table}[t]
\centering
\caption{Prediction correlation: full fine-tuning re-converges, low-rank isolation does not.}
\label{tab:decorr}
\footnotesize\setlength{\tabcolsep}{4pt}\begin{tabular}{@{}lc@{}}
\toprule
Member pair (all \Dtwo{} \textsc{expr}) & pred.\ correlation $r$ \\
\midrule
Full-FT curricula, one backbone & 0.98 \\
LoRA experts, one backbone & 0.91 \\
AffectNet-LoRA vs.\ a different backbone & 0.75 \\
FSFM vs.\ MAE-Face (two backbones) & 0.81 \\
\bottomrule
\end{tabular}

\end{table}

\section{Conclusion}
\label{sec:conclusion}
Ensemble construction on this benchmark proceeds one member at a time, and the operative question is which
member to add. We answered it with a strength-parity rule, requiring an added member to be both
decorrelated and a near-peer, and with the mechanism the rule needs, parameter-isolation, which
manufactures decorrelated near-peers from a single backbone where full fine-tuning cannot. On the 11th ABAW
MTL track the resulting system reaches \arxThreeBBPMTL{} on validation and \arxVApoolPMTL{} in its strongest
configuration. The rule also implies its own limit: once the pool holds a few decorrelated near-peers, any
further member is almost always correlated or weaker and the score plateaus, so closing the remaining gap
will depend on a genuinely different near-peer representation rather than on new ways of averaging the
existing members.

\bibliographystyle{IEEEtran}
\bibliography{refs}

@inproceedings{abaw2022augraph,
  title={Affective Behavior Analysis using Action Unit Relation Graph and Multi-task Cross Attention},
  author={Nguyen, Dang-Khanh and Pant, Sudarshan and Ho, Ngoc-Huynh and Lee, Guee-Sang and Kim, Soo-Hyung and Yang, Hyung-Jeong},
  booktitle={Computer Vision -- ECCV 2022 Workshops}, series={Lecture Notes in Computer Science},
  pages={132--142}, year={2023}, publisher={Springer}, doi={10.1007/978-3-031-25075-0\_10}}

@inproceedings{abaw2024auassisted,
  title={Affective Behavior Analysis using Task-adaptive and {AU}-assisted Graph Network},
  author={Li, Xiaodong and Du, Wenchao and Yang, Hongyu},
  booktitle={Computer Vision -- ECCV 2024 Workshops}, series={Lecture Notes in Computer Science},
  pages={393--403}, year={2025}, publisher={Springer}, doi={10.1007/978-3-031-91581-9\_28}}

@inproceedings{abaw2024progressive,
  title={Affective Behaviour Analysis via Progressive Learning},
  author={Liu, Chen and Zhang, Wei and Qiu, Feng and Li, Lincheng and Wang, Dadong and Yu, Xin},
  booktitle={Computer Vision -- ECCV 2024 Workshops}, series={Lecture Notes in Computer Science},
  pages={366--379}, year={2025}, publisher={Springer}, doi={10.1007/978-3-031-91581-9\_26}}

@article{kollias2019deep,
  title={Deep affect prediction in-the-wild: Aff-wild database and challenge, deep architectures, and beyond},
  author={Kollias, Dimitrios and Tzirakis, Panagiotis and Nicolaou, Mihalis A and Papaioannou, Athanasios and Zhao, Guoying and Schuller, Bj{\"o}rn and Kotsia, Irene and Zafeiriou, Stefanos},
  journal={International Journal of Computer Vision}, pages={1--23}, year={2019}, publisher={Springer},
  doi={10.1007/s11263-019-01158-4}}

@article{kollias2019expression,
  title={Expression, Affect, Action Unit Recognition: Aff-Wild2, Multi-Task Learning and ArcFace},
  author={Kollias, Dimitrios and Zafeiriou, Stefanos},
  journal={arXiv preprint arXiv:1910.04855}, year={2019}}

@article{kollias2019face,
  title={Face Behavior a la carte: Expressions, Affect and Action Units in a Single Network},
  author={Kollias, Dimitrios and Sharmanska, Viktoriia and Zafeiriou, Stefanos},
  journal={arXiv preprint arXiv:1910.11111}, year={2019}}

@inproceedings{kollias2020analysing,
  title={Analysing Affective Behavior in the First ABAW 2020 Competition},
  author={Kollias, D and Schulc, A and Hajiyev, E and Zafeiriou, S},
  booktitle={2020 15th IEEE International Conference on Automatic Face and Gesture Recognition (FG 2020)}, pages={794--800}, year={2020}, doi={10.1109/FG47880.2020.00126}}

@article{kollias2021affect,
  title={Affect Analysis in-the-wild: Valence-Arousal, Expressions, Action Units and a Unified Framework},
  author={Kollias, Dimitrios and Zafeiriou, Stefanos},
  journal={arXiv preprint arXiv:2103.15792}, year={2021}}

@article{kollias2021distribution,
  title={Distribution Matching for Heterogeneous Multi-Task Learning: a Large-scale Face Study},
  author={Kollias, Dimitrios and Sharmanska, Viktoriia and Zafeiriou, Stefanos},
  journal={arXiv preprint arXiv:2105.03790}, year={2021}}

@inproceedings{kollias2021analysing,
  title={Analysing affective behavior in the second abaw2 competition},
  author={{Kollias}, Dimitrios and {Zafeiriou}, Stefanos},
  booktitle={Proceedings of the IEEE/CVF International Conference on Computer Vision}, pages={3652--3660}, year={2021},
  doi={10.1109/ICCVW54120.2021.00408}}

@inproceedings{kollias2022abaw,
  title={Abaw: Valence-arousal estimation, expression recognition, action unit detection \& multi-task learning challenges},
  author={Kollias, Dimitrios},
  booktitle={Proceedings of the IEEE/CVF Conference on Computer Vision and Pattern Recognition}, pages={2328--2336}, year={2022},
  doi={10.1109/CVPRW56347.2022.00259}}

@inproceedings{kollias2023abaw,
  title={Abaw: Learning from synthetic data \& multi-task learning challenges},
  author={{Kollias}, Dimitrios},
  booktitle={European Conference on Computer Vision}, pages={157--172}, year={2023}, organization={Springer},
  doi={10.1007/978-3-031-25075-0\_12}}

@inproceedings{kollias2023abaw2,
  title={Abaw: Valence-arousal estimation, expression recognition, action unit detection \& emotional reaction intensity estimation challenges},
  author={Kollias, Dimitrios and Tzirakis, Panagiotis and Baird, Alice and Cowen, Alan and Zafeiriou, Stefanos},
  booktitle={Proceedings of the IEEE/CVF Conference on Computer Vision and Pattern Recognition}, pages={5888--5897}, year={2023},
  doi={10.1109/CVPRW59228.2023.00626}}

@inproceedings{kollias20246th,
  title={The 6th affective behavior analysis in-the-wild (abaw) competition},
  author={Kollias, Dimitrios and Tzirakis, Panagiotis and Cowen, Alan and Zafeiriou, Stefanos and Kotsia, Irene and Baird, Alice and Gagne, Chris and Shao, Chunchang and Hu, Guanyu},
  booktitle={Proceedings of the IEEE/CVF Conference on Computer Vision and Pattern Recognition}, pages={4587--4598}, year={2024},
  doi={10.1109/CVPRW63382.2024.00461}}

@inproceedings{kollias2024abaw7,
  title={7th abaw competition: Multi-task learning and compound expression recognition},
  author={Kollias, Dimitrios and Zafeiriou, Stefanos and Kotsia, Irene and Dhall, Abhinav and Ghosh, Shreya and Shao, Chunchang and Hu, Guanyu},
  booktitle={European Conference on Computer Vision}, pages={31--45}, year={2024}, organization={Springer},
  doi={10.1007/978-3-031-91581-9\_3}}

@article{kollias2024behaviour4all,
  title={Behaviour4all: in-the-wild facial behaviour analysis toolkit},
  author={Kollias, Dimitrios and Shao, Chunchang and Kaloidas, Odysseus and Patras, Ioannis},
  journal={arXiv preprint arXiv:2409.17717}, year={2024}}

@inproceedings{kollias2024distribution,
  title={Distribution matching for multi-task learning of classification tasks: a large-scale study on faces \& beyond},
  author={Kollias, Dimitrios and Sharmanska, Viktoriia and Zafeiriou, Stefanos},
  booktitle={Proceedings of the AAAI Conference on Artificial Intelligence},
  volume={38}, number={3}, pages={2813--2821}, year={2024}, doi={10.1609/aaai.v38i3.28061}}

@inproceedings{kollias2025advancements,
  title={Advancements in Affective and Behavior Analysis: The 8th ABAW Workshop and Competition},
  author={Kollias, Dimitrios and Tzirakis, Panagiotis and Cowen, Alan and Zafeiriou, Stefanos and Kotsia, Irene and Granger, Eric and Pedersoli, Marco and Bacon, Simon and Baird, Alice and Gagne, Chris and others},
  booktitle={Proceedings of the Computer Vision and Pattern Recognition Conference}, pages={5572--5583}, year={2025},
  doi={10.1109/CVPRW67362.2025.00554}}

@inproceedings{kollias2025emotions,
  title={From emotions to violence: Multimodal fine-grained behavior analysis at the 9th abaw},
  author={Kollias, Dimitrios and Zafeiriou, Stefanos and Kotsia, Irene and Slabaugh, Greg and Senadeera, Damith Chamalke and Zheng, Jianian and Yadav, Kaushal Kumar Keshlal and Shao, Chunchang and Hu, Guanyu},
  booktitle={Proceedings of the IEEE/CVF International Conference on Computer Vision}, pages={1--12}, year={2025},
  doi={10.1109/ICCVW69036.2025.00006}}

@misc{kollias2026affect,
  title={From Affect to Complex Behavior: Advancing Multimodal Human-Centered AI at the 10th ABAW Workshop \& Competition},
      author    = {Kollias, Dimitrios and Tzirakis, Panagiotis and Cowen, Alan and Zafeiriou, Stefanos and Kotsia, Irene and Granger, Eric and Pedersoli, Marco and Bacon, Simon and Madsen, Jens and Belharbi, Soufiane and Aslam, Muhammad Haseeb and Shao, Chunchang and Hu, Guanyu},
  year={2026}, eprint={2605.27451}, archivePrefix={arXiv}}

@article{savchenko2024hsemotion,
  title={{HSEmotion} Team at the 7th {ABAW} Challenge: Multi-Task Learning and Compound Facial Expression Recognition},
  author={Savchenko, Andrey V.}, journal={arXiv preprint arXiv:2407.13184}, year={2024}}

@inproceedings{wang2025fsfm,
  title={{FSFM}: A Generalizable Face Security Foundation Model via Self-Supervised Facial Representation Learning},
  author={Wang, Gaojian and Lin, Feng and Wu, Tong and Liu, Zhenguang and Ba, Zhongjie and Ren, Kui},
  booktitle={IEEE/CVF Conference on Computer Vision and Pattern Recognition (CVPR)},
  year={2025}, doi={10.1109/CVPR52734.2025.02269}}

@inproceedings{zafeiriou2017aff,
  title={Aff-wild: Valence and arousal `in-the-wild' challenge},
  author={Zafeiriou, Stefanos and Kollias, Dimitrios and Nicolaou, Mihalis A and Papaioannou, Athanasios and Zhao, Guoying and Kotsia, Irene},
  booktitle={Computer Vision and Pattern Recognition Workshops (CVPRW), 2017 IEEE Conference on}, pages={1980--1987}, year={2017}, organization={IEEE}}

@misc{companion,
  title={A Shared Latent for Partially-Labeled Multi-Task Facial Affect Recognition},
  author={Hong Hai Nguyen and Sy Phan Van and Soo-Hyung Kim and Van-Thong Huynh},
  year={2026},
  eprint={2607.16285},
  archivePrefix={arXiv},
  primaryClass={cs.CV},
  url={https://arxiv.org/abs/2607.16285},
  }

@inproceedings{kingma2014semi,
  title={Semi-Supervised Learning with Deep Generative Models},
  author={Kingma, Diederik P. and Rezende, Danilo J. and Mohamed, Shakir and Welling, Max},
  booktitle={Advances in Neural Information Processing Systems (NeurIPS)},
  volume={27}, year={2014}}

@inproceedings{ma2023maeface,
  title={Multi-modal Facial Affective Analysis based on Masked Autoencoder},
  author={Zhang, Wei and Ma, Bowen and Qiu, Feng and Ding, Yu},
  booktitle={IEEE/CVF Conference on Computer Vision and Pattern Recognition Workshops (CVPRW)},
  year={2023},
  }

@misc{wang2021meanteacher,
  title={A Multi-task Mean Teacher for Semi-supervised Facial Affective Behavior Analysis},
  author={Wang, Lingfeng and Wang, Shisen and Qi, Jin and Suzuki, Kenji},
  year={2021}, eprint={2107.04225}, archivePrefix={arXiv}, primaryClass={cs.CV},
  doi={10.48550/arXiv.2107.04225},
  }

@misc{gera2022ssmfar,
  title={{SS-MFAR}: Semi-supervised Multi-task Facial Affect Recognition},
  author={Gera, Darshan and Kumar, Badveeti Naveen Siva and Kumar, Bobbili Veerendra Raj and Balasubramanian, S.},
  year={2022}, eprint={2207.09012}, archivePrefix={arXiv}, primaryClass={cs.CV},
  doi={10.48550/arXiv.2207.09012},
  }

@article{mollahosseini2019affectnet,
  title={{AffectNet}: A Database for Facial Expression, Valence, and Arousal Computing in the Wild},
  author={Mollahosseini, Ali and Hasani, Behzad and Mahoor, Mohammad H.},
  journal={IEEE Transactions on Affective Computing}, volume={10}, number={1}, pages={18--31},
  year={2019}, publisher={IEEE}, doi={10.1109/TAFFC.2017.2740923}}

@inproceedings{hu2022lora,
  title={{LoRA}: Low-Rank Adaptation of Large Language Models},
  author={Hu, Edward J. and Shen, Yelong and Wallis, Phillip and Allen-Zhu, Zeyuan and Li, Yuanzhi
         and Wang, Shean and Wang, Lu and Chen, Weizhu},
  booktitle={International Conference on Learning Representations (ICLR)}, year={2022}}

@article{oquab2024dinov2,
  title={{DINOv2}: Learning Robust Visual Features without Supervision},
  author={Oquab, Maxime and Darcet, Timoth{\'e}e and Moutakanni, Th{\'e}o and others},
  journal={Transactions on Machine Learning Research (TMLR)}, year={2024}}

@inproceedings{krogh1995ensembles,
  title={Neural Network Ensembles, Cross Validation, and Active Learning},
  author={Krogh, Anders and Vedelsby, Jesper},
  booktitle={Advances in Neural Information Processing Systems (NIPS)},
  volume={7}, pages={231--238}, year={1995},
  }

@article{brown2005diversity,
  title={Diversity creation methods: a survey and categorisation},
  author={Brown, Gavin and Wyatt, Jeremy and Harris, Rachel and Yao, Xin},
  journal={Information Fusion}, volume={6}, number={1}, pages={5--20}, year={2005},
  doi={10.1016/j.inffus.2004.04.004}}

@inproceedings{lakshminarayanan2017deep,
  title={Simple and Scalable Predictive Uncertainty Estimation using Deep Ensembles},
  author={Lakshminarayanan, Balaji and Pritzel, Alexander and Blundell, Charles},
  booktitle={Advances in Neural Information Processing Systems (NeurIPS)}, year={2017},
  doi={10.48550/arXiv.1612.01474}}

@article{wang2023loraens,
  title={{LoRA} Ensembles for Large Language Model Fine-Tuning},
  author={Wang, Xi and Aitchison, Laurence and Rudolph, Maja},
  journal={arXiv preprint arXiv:2310.00035}, year={2023},
  doi={10.48550/arXiv.2310.00035}}

@inproceedings{benitezquiroz2016emotionet,
  title={{EmotioNet}: An Accurate, Real-Time Algorithm for the Automatic Annotation of a Million Facial Expressions in the Wild},
  author={Benitez-Quiroz, C. Fabian and Srinivasan, Ramprakash and Martinez, Aleix M.},
  booktitle={IEEE Conf. Computer Vision and Pattern Recognition (CVPR)},
  pages={5562--5570}, year={2016}, doi={10.1109/CVPR.2016.600}}
\end{document}